\pdfoutput=1
\documentclass[11pt]{article}
\usepackage[table]{xcolor}

\usepackage[final]{acl}

\usepackage{times}
\usepackage{latexsym}

\usepackage[T1]{fontenc}
\usepackage[utf8]{inputenc}

\usepackage{microtype}

\usepackage{inconsolata}

\usepackage{graphicx}
\usepackage{subcaption}
\usepackage{array}
\usepackage{booktabs}
\usepackage{makecell}
\usepackage{subcaption} % Add this line for subcaptions
\usepackage{amsmath}

\usepackage{highlight}
\renewcommand{\opacity}{50}

\title{Language Models Do Hard Arithmetic Tasks Easily and Hardly Do Easy Arithmetic Tasks}

\author{
 \textbf{Andrew Gambardella\thanks{Correspondence: \href{mailto:atgambardella@weblab.t.u-tokyo.ac.jp}{atgambardella@weblab.t.u-tokyo.ac.jp}}} \quad
 \textbf{Yusuke Iwasawa} \quad
 \textbf{Yutaka Matsuo}
\\
 University of Tokyo
\\
}

\begin{document}
\maketitle
\begin{abstract}
The ability (and inability) of large language models (LLMs) to perform arithmetic tasks has been the subject of much theoretical and practical debate.
We show that LLMs are frequently able to correctly and confidently predict the first digit of $n$-digit by $m$-digit multiplication tasks without using chain of thought reasoning, despite these tasks require compounding operations to solve.
Simultaneously, LLMs in practice often fail to correctly or confidently predict the last digit of an $n$-digit by $m$-digit multiplication, a task equivalent to $1$-digit by $1$-digit multiplication which can be easily learned or memorized.
We show that the latter task can be solved more robustly when the LLM is conditioned on all of the correct higher-order digits, which on average increases the confidence of the correct last digit on $5$-digit by $5$-digit multiplication tasks using Llama 2-13B by over 230\% (0.13→0.43) and Mistral-7B by 150\% (0.22→0.55).\end{abstract}

\section{Introduction}

The development of large language models (LLMs)~\cite{brown2020language} has given new life to the deep learning revolution, and seen mass adoption within not just the scientific community, but also society at large.
These LLMs, being the first known ``general'' machine learning model developed by humanity~\cite{morris2023agilevels}, have been applied to various tasks dealing with natural language such as those commonly encountered in school curricula~\cite{Hendrycks2021MMLU}, and even branching off into tasks such as text-to-image generation~\cite{saharia2022imagen} and hierarchical planning~\cite{wang2023voyager}.

Despite the generality and far-reaching consequences of LLMs, there are still many significant limitations making difficult the direct application of LLMs to certain tasks.
One such limitation is the poor performance of LLMs on arithmetic tasks, such as elementary addition, subtraction, multiplication, and division~\cite{nogueira2021arithmetic}.
Not only do modern LLMs perform poorly on these tasks, but some tasks such as $n$-digit by $m$-digit multiplication and division, which require compounding operations to solve, appear to be unlearnable by pure autoregressive transformer architectures unless they decompose the problem into multiple steps, such as with chain of thought reasoning~\cite{wies2022sub,liu2023goat}.
As such, several solutions have been proposed, such as fine-tuning so that chain of thought reasoning is automatically used for problems which require compounding operations~\cite{liu2023goat,kojima2022large} or fine-tuning to call outside tools, such as a calculator~\cite{schick2024toolformer}.

While we most likely cannot expect simply training models with more parameters to allow for the solving of tasks which require compounding operations without chain of thought, we believe that analyzing the limitations and abilities of autoregressive LLMs when attempting to solve these tasks directly may shed light on unknown properties of LLMs.
We therefore use Monte Carlo Dropout (MC Dropout)~\cite{Gal2015} to analyze the performance of LLMs which were trained with dropout and which have open weights available, such as Llama 2~\cite{touvron2023llama} and Mistral~\cite{jiang2023mistral}, in carrying out arithmetic tasks.

MC Dropout allows one to interpret neural networks which were trained with dropout as Bayesian neural networks, as neural networks trained with dropout have been shown to be equivalent to a Bayesian approximation to a Gaussian process. This allows one to obtain empirical Bayesian confidence distributions over neural network weights or outputs by doing multiple forward passes through the neural network with dropout on, during test time~\cite{Gal2015}. MC Dropout is one of many ensemble-based methods for uncertainty quantification~\cite{ovadia2019uncertainty, ashukha2020pitfalls}, and has been applied to analyze the confidence of transformer architectures~\cite{shelmanov2021certain} and to implement tree-based LLM prompting~\cite{mo2023tout}.

Our results when applying MC Dropout to Llama 2 and Mistral in arithmetic tasks were surprising. We found that all models could confidently and correctly predict the first digit result of $n$-digit by $m$-digit multiplication problems, despite it most likely being impossible for any autoregressive LLM to have learned a general algorithm for doing so without decomposing the problem into multiple steps, as finding this digit in general requires solving the entire multiplication problem\footnote{Consider that the highest-order digit of $31622776601683793319^2$ is $9$, but the highest-order digit of $31622776601683793320^2$ is $1$.}.
We also found that all models struggled to correctly output the last digit of $n$-digit by $m$-digit multiplication problems, despite it being very easy to learn an algorithm for doing so, as calculating the last digit is equivalent to $1$-digit by $1$-digit multiplication.
Finally, we show that the confidence of LLMs in predicting the last digit can be increased by conditioning the generation of the last digit on the correct intervening digits, despite the computation of the last digit not depending on the correct computations of the higher-order digits at all.

\section{Experiments}
\label{sec:experiments}

We evaluate the HuggingFace~\cite{wolf2019huggingface} implementations of Llama 2-7B, Llama 2-13B, and Mistral-7B~\cite{touvron2023llama, jiang2023mistral} in 2-shot settings, where the 2-shot examples are of correct $n$-digit by $m$-digit multiplications.
Sections~\ref{sec:unconditional} and~\ref{sec:conditional} show results on the $3$-digit by $3$-digit multiplication task $592*392$, and averages over multiple problems with varying digit length are provided in Section~\ref{sec:many}.
Details about the prompt and hyperparameters are given in Appendix~\ref{sec:appendixHyperparam}, details about the tokenizers for the models are given in Appendix~\ref{sec:appendixTokenization}, and details about the use of dropout in the training of the models is given in Appendix~\ref{sec:appendixDropout}.

\subsection{Unconditional Answer Generation}
\label{sec:unconditional}

\begin{figure}[h]
    \centering

    \begin{subfigure}{0.44\textwidth}
        \includegraphics[width=\linewidth]{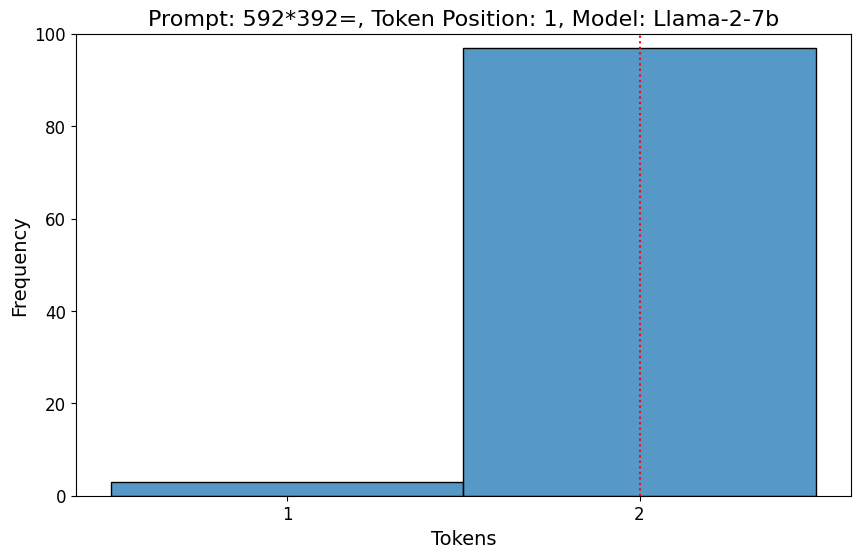}
    \end{subfigure}
    \hfill
    \begin{subfigure}{0.44\textwidth}
        \includegraphics[width=\linewidth]{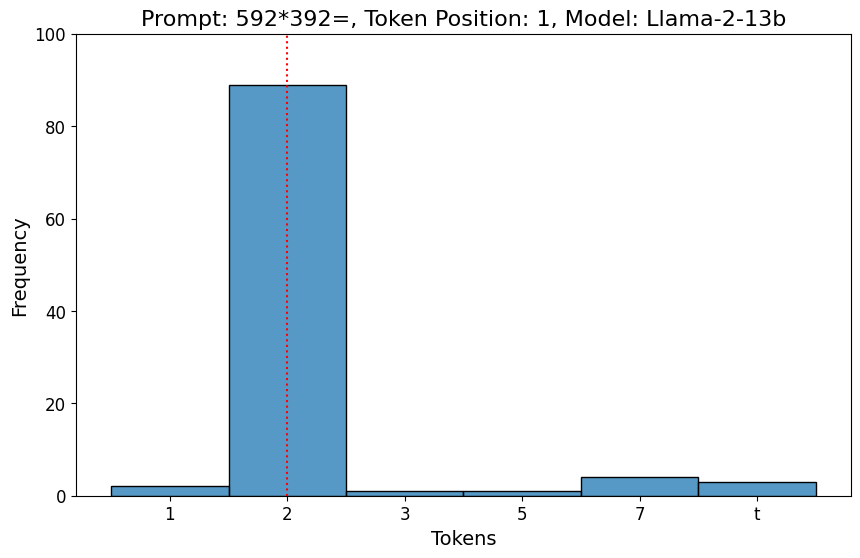}
    \end{subfigure}

    \caption{Confidence and accuracy of Llama 2-7B and Llama 2-13B predicting the first digit of the result of $592*392$.
    Both language models are able to confidently and correctly predict that the first digit should be $2$, despite this not being immediately apparent from the problem.
    }
    \label{fig:first_digit_unconditional}
\end{figure}

\begin{figure}[h]
    \centering

    \begin{subfigure}{0.44\textwidth}
        \includegraphics[width=\linewidth]{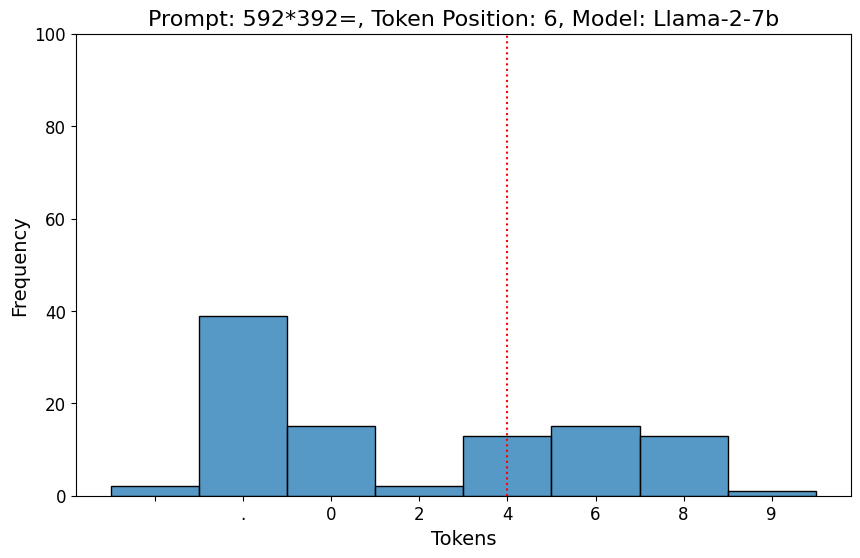}
    \end{subfigure}
    \hfill
    \begin{subfigure}{0.44\textwidth}
        \includegraphics[width=\linewidth]{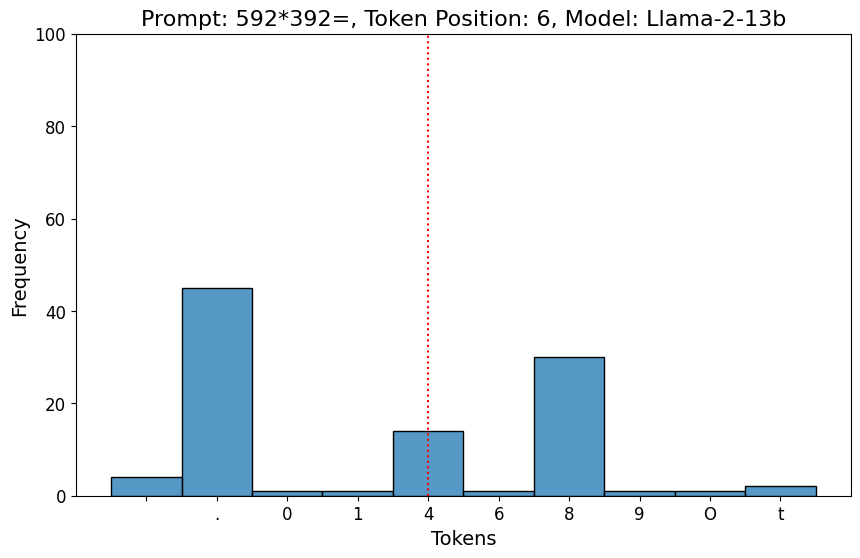}
    \end{subfigure}

    \caption{Confidence and accuracy of Llama 2-7B and Llama 2-13B predicting the sixth digit of the result of $592*392$.
    Neither are able to predict this digit confidently, with the mode of the distribution on the ``end string'' character in both cases.
    Both only output $4$ in about 20\% of samples, despite it being immediately apparent that the final digit should be $4$.}
    \label{fig:last_digit_unconditional}
\end{figure}

We first study a version of the problem in which the answer is generated with the language model conditioned on the few shot examples and the problem to be solved, but is provided with none of the digits to be generated (i.e., the normal few-shot arithmetic scenario), which we refer to as ``unconditional'' generation in an abuse of terminology. Our main results for these experiments are in Figures~\ref{fig:first_digit_unconditional} and~\ref{fig:last_digit_unconditional}.

In Figure~\ref{fig:first_digit_unconditional} we can see that both Llama 2-7B and Llama 2-13B can confidently and correctly predict the first digit of the $3$-digit by $3$-digit multiplication task $592*392$, which equals $232064$. This should be surprising as it is not immediately apparent from the problem that the first digit of the solution should be $2$, and the only way to discover this is to compute the multiplication. As LLMs most likely cannot perform $n$-digit by $m$-digit multiplication in the general case without decomposing the problem into steps, the output of the first digit in this case is unlikely to be the output of a multiplication algorithm learned by the LLM.

Conversely, in Figure~\ref{fig:last_digit_unconditional}, we can see that both Llama 2-7B and Llama 2-13B can neither confidently nor correctly predict the last digit of the same problem, despite doing so being equivalent to $1$-digit by $1$-digit multiplication. This is a case in which any reasonable model should be able to confidently and correctly solve the task, as not only could the algorithm to solve the task be learned by an autoregressive language model, but the information needed to solve this task could also very easily be memorized by language models with billions of weights.

\subsection{Conditional Answer Generation}
\label{sec:conditional}

\begin{figure}[h]
    \centering

    \begin{subfigure}{0.44\textwidth}
        \includegraphics[width=\linewidth]{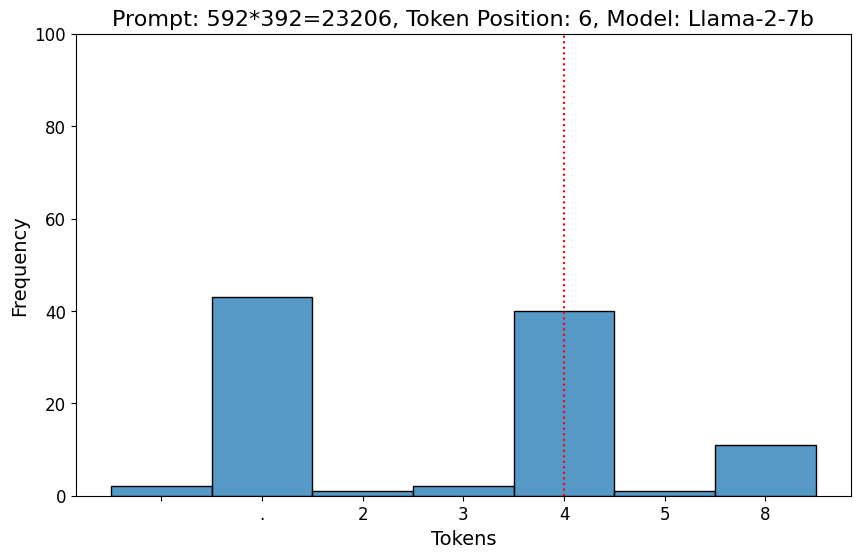}
    \end{subfigure}
    \hfill
    \begin{subfigure}{0.44\textwidth}
        \includegraphics[width=\linewidth]{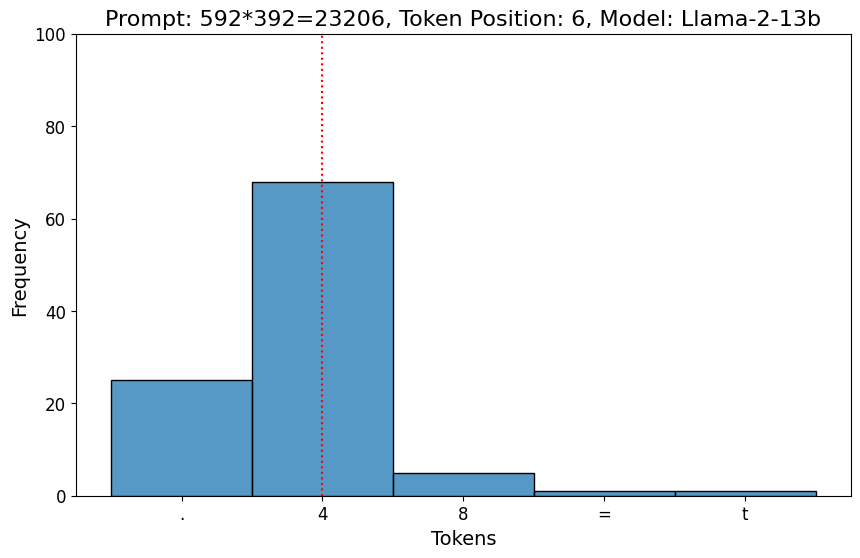}
    \end{subfigure}

    \caption{Confidence and accuracy of Llama 2-7B and Llama 2-13B predicting the last digit of the result of $592*392$, when conditioned on the first five correct digits.
    The confidence in the correct answer being $4$ doubles for Llama 2-7B and more than triples for Llama 2-13B, despite the computation of the last digit not depending on the prior digits being correct at all.
    }
    \label{fig:last_digit_conditional}
\end{figure}

Finally, we contrast the experiments given in Figures~\ref{fig:first_digit_unconditional} and~\ref{fig:last_digit_unconditional} with a third experiment, in which the LLM is given all digits from the answer except for the final digit, and is tasked with outputting solely the final digit, which we refer to as ``conditional'' generation in an abuse of terminology.
Results for this experiment are given in Figure~\ref{fig:last_digit_conditional}.
In this case the confidence in the correct output doubles for Llama 2-7B and triples for Llama 2-13B, with Llama 2-13B now having most of its probability mass on the correct last digit, whereas it did not do so when generating the entire string at once (and therefore often conditioning on incorrect prior digits).
The fact that in both cases, more probability mass is being put on the correct answer should be surprising, as the computation of this digit does not depend on the correctness of the higher-order digits in any way.

\subsection{Ablation Over Digit Length}
\label{sec:many}

We provide further ablations over digit length with Llama 2-7B and 13B in Table~\ref{tab:llama2results}. Each subtable gives the confidence of the correct digit, averaged over 10 different $n$-digit by $m$-digit multiplication problems each.
We find that the conclusions shown for a single example in Sections~\ref{sec:unconditional} and \ref{sec:conditional} hold over varying multiplication problems and digit lengths in general. We further provide similar Mistral-7B experiments in Table~\ref{tab:mistralresults}. While Mistral-7B is stronger at arithmetic tasks than both Llama 2-7B and 13B, the same patterns and conclusions found for Llama 2-7B and 13B also hold for Mistral-7B.

\begin{table*}[ht]
\centering
\begin{tabular}{ cc }
Llama 2-7B & Llama 2-13B \\

    \begin{subtable}{0.49\linewidth} % Subtable (a)
    \centering
    \begin{tabular}{|c|c|c|c|c|}
        \hline
        \diaghead{\theadfont Diaga a a a II}%
        {n}{m} & 2 & 3 & 4 & 5 \\
        \hline
        2 & \gradient{0.81} & \gradient{0.90} & \gradient{0.82} & \gradient{0.82} \\
        \hline
        3 & \gradient{0.91} & \gradient{0.78} & \gradient{0.88} & \gradient{0.92} \\
        \hline
        4 & \gradient{0.88} & \gradient{0.83} & \gradient{0.92} & \gradient{0.77} \\
        \hline
        5 & \gradient{0.89} & \gradient{0.74} & \gradient{0.89} & \gradient{0.87} \\
        \hline
    \end{tabular}
    \caption{}
    \label{llamasubtab:a}
    \end{subtable}
    &

    \begin{subtable}{0.49\linewidth} % Subtable (b)
    \centering
    \begin{tabular}{|c|c|c|c|c|}
        \hline
        \diaghead{\theadfont Diaga a a a II}%
        {n}{m} & 2 & 3 & 4 & 5 \\
        \hline
        2 & \gradient{0.84} & \gradient{0.85} & \gradient{0.79} & \gradient{0.73} \\
        \hline
        3 & \gradient{0.87} & \gradient{0.72} & \gradient{0.85} & \gradient{0.86} \\
        \hline
        4 & \gradient{0.84} & \gradient{0.83} & \gradient{0.78} & \gradient{0.78} \\
        \hline
        5 & \gradient{0.86} & \gradient{0.71} & \gradient{0.84} & \gradient{0.86} \\
        \hline
    \end{tabular}
    \caption{}
    \label{llamasubtab:b}
    \end{subtable}
    \\

    \begin{subtable}{0.49\linewidth} % Subtable (c)
    \centering
    \begin{tabular}{|c|c|c|c|c|}
        \hline
        \diaghead{\theadfont Diaga a a a II}%
        {n}{m} & 2 & 3 & 4 & 5 \\
        \hline
        2 & \gradient{0.52} & \gradient{0.34} & \gradient{0.16} & \gradient{0.20} \\
        \hline
        3 & \gradient{0.39} & \gradient{0.22} & \gradient{0.16} & \gradient{0.19} \\
        \hline
        4 & \gradient{0.40} & \gradient{0.21} & \gradient{0.20} & \gradient{0.15} \\
        \hline
        5 & \gradient{0.33} & \gradient{0.20} & \gradient{0.15} & \gradient{0.11} \\
        \hline
    \end{tabular}
    \caption{}
    \label{llamasubtab:c}
    \end{subtable}
    &

    \begin{subtable}{0.49\linewidth} % Subtable (d)
    \centering
    \begin{tabular}{|c|c|c|c|c|}
        \hline
        \diaghead{\theadfont Diaga a a a II}%
        {n}{m} & 2 & 3 & 4 & 5 \\
        \hline
        2 & \gradient{0.78} & \gradient{0.50} & \gradient{0.32} & \gradient{0.30} \\
        \hline
        3 & \gradient{0.56} & \gradient{0.40} & \gradient{0.24} & \gradient{0.17} \\
        \hline
        4 & \gradient{0.63} & \gradient{0.37} & \gradient{0.29} & \gradient{0.22} \\
        \hline
        5 & \gradient{0.52} & \gradient{0.30} & \gradient{0.24} & \gradient{0.13} \\
        \hline
    \end{tabular}
    \caption{}
    \label{llamasubtab:d}
    \end{subtable} \\

    \begin{subtable}{0.49\linewidth} % Subtable (e)
    \centering
    \begin{tabular}{|c|c|c|c|c|}
        \hline
        \diaghead{\theadfont Diaga a a a II}%
        {n}{m} & 2 & 3 & 4 & 5 \\
        \hline
        2 & \gradient{0.64} & \gradient{0.41} & \gradient{0.24} & \gradient{0.51} \\
        \hline
        3 & \gradient{0.55} & \gradient{0.45} & \gradient{0.38} & \gradient{0.40} \\
        \hline
        4 & \gradient{0.43} & \gradient{0.33} & \gradient{0.38} & \gradient{0.36} \\
        \hline
        5 & \gradient{0.44} & \gradient{0.41} & \gradient{0.26} & \gradient{0.25} \\
        \hline
    \end{tabular}
    \caption{}
    \label{llamasubtab:e}
    \end{subtable}
    &

    \begin{subtable}{0.49\linewidth} % Subtable (f)
    \centering
    \begin{tabular}{|c|c|c|c|c|}
        \hline
        \diaghead{\theadfont Diaga a a a II}%
        {n}{m} & 2 & 3 & 4 & 5 \\
        \hline
        2 & \gradient{0.82} & \gradient{0.66} & \gradient{0.48} & \gradient{0.57} \\
        \hline
        3 & \gradient{0.66} & \gradient{0.68} & \gradient{0.49} & \gradient{0.51} \\
        \hline
        4 & \gradient{0.73} & \gradient{0.54} & \gradient{0.56} & \gradient{0.47} \\
        \hline
        5 & \gradient{0.70} & \gradient{0.54} & \gradient{0.50} & \gradient{0.43} \\
        \hline
    \end{tabular}
    \caption{}
    \label{llamasubtab:f}
    \end{subtable} \\

\end{tabular}
    \caption{Llama 2-7B and 13B generation average confidence of the correct first digit (a, b), unconditional average confidence of the correct last digit (c, d), and conditional average confidence of the correct last digit (e, f).}
    \label{tab:llama2results}
\end{table*}

\begin{table*}[h!t]
\centering
    \begin{subtable}{0.99\linewidth} % Subtable (a)
    \centering

    \begin{tabular}{|c|c|c|c|c|}
        \hline
        \diaghead{\theadfont Diaga a a a II}%
        {n}{m} & 2 & 3 & 4 & 5 \\
        \hline
2 & \gradient{0.97}± 0.03 & \gradient{0.98}± 0.03 & \gradient{0.98}± 0.02 & \gradient{1.00}± 0.00 \\
\hline
3 & \gradient{0.98}± 0.03 & \gradient{1.00}± 0.00 & \gradient{0.94}± 0.09 & \gradient{0.93}± 0.04 \\
\hline
4 & \gradient{0.99}± 0.01 & \gradient{0.87}± 0.15 & \gradient{0.98}± 0.04 & \gradient{0.82}± 0.09 \\
\hline
5 & \gradient{0.89}± 0.1 & \gradient{0.94}± 0.11 & \gradient{0.95}± 0.06 & \gradient{0.99}± 0.01 \\
\hline

    \end{tabular}
    \caption{}
    \label{mistralsubtab:a}
    \end{subtable}
    \\

    \begin{subtable}{0.99\linewidth} % Subtable (b)
    \centering

    \begin{tabular}{|c|c|c|c|c|}
        \hline
        \diaghead{\theadfont Diaga a a a II}%
        {n}{m} & 2 & 3 & 4 & 5 \\
        \hline
2 & \gradient{0.74}± 0.06 & \gradient{0.57}± 0.26 & \gradient{0.52}± 0.29 & \gradient{0.41}± 0.21 \\
\hline
3 & \gradient{0.87}± 0.10 & \gradient{0.70}± 0.13 & \gradient{0.20}± 0.12 & \gradient{0.11}± 0.07 \\
\hline
4 & \gradient{0.44}± 0.14 & \gradient{0.70}± 0.14 & \gradient{0.28}± 0.23 & \gradient{0.30}± 0.15 \\
\hline
5 & \gradient{0.70}± 0.10 & \gradient{0.33}± 0.09 & \gradient{0.20}± 0.13 & \gradient{0.22}± 0.07 \\
\hline

    \end{tabular}
    \caption{}
    \label{mistralsubtab:b}
    \end{subtable}
    \\

    \begin{subtable}{0.99\linewidth} % Subtable (c)
    \centering
    \begin{tabular}{|c|c|c|c|c|}
        \hline
        \diaghead{\theadfont Diaga a a a II}%
        {n}{m} & 2 & 3 & 4 & 5 \\
        \hline
2 & \gradient{0.85}± 0.23 & \gradient{0.83}± 0.13 & \gradient{0.73}± 0.21 & \gradient{0.76}± 0.23 \\
\hline
3 & \gradient{0.86}± 0.13 & \gradient{0.85}± 0.11 & \gradient{0.75}± 0.22 & \gradient{0.57}± 0.32 \\
\hline
4 & \gradient{0.76}± 0.17 & \gradient{0.62}± 0.27 & \gradient{0.77}± 0.26 & \gradient{0.59}± 0.26 \\
\hline
5 & \gradient{0.80}± 0.18 & \gradient{0.68}± 0.21 & \gradient{0.65}± 0.26 & \gradient{0.55}± 0.35 \\
\hline

    \end{tabular}
    \caption{}
    \label{mistralsubtab:c}
    \end{subtable}
    \\
\caption{Mistral-7B generation average and standard deviation confidence of the correct first digit (a), unconditional average and standard deviation confidence of the correct last digit (b), and conditional average and standard deviation confidence of the correct last digit (c).}
\label{tab:mistralresults}
\end{table*}

\section{Discussion of Results}

\subsection{First Digit}

It is most likely impossible for autoregressive LLMs to compute the first digit of an $n$-digit by $m$-digit multiplication problem without decomposing the problem into steps, especially given that the answer is being written starting with the highest-order digit, and calculating the first digit depends on the correct calculations of the lower-order digits.

LLMs \emph{can}, however, perform $1$-digit by $1$-digit multiplication. If these LLMs were to internally round $592$ to $600$ and $392$ to $400$, it could approximately solve for the highest-order digit in this way, as $600*400$ is a computation that can be performed by autoregressive language models. We find it likely that such a computation is occurring inside these LLMs, especially as stochastic gradient descent is likely to find such ``shortcuts.'' 

\subsection{Last Digit}

Both LLMs failing to predict the last digit when generating the entire string autoregressively, and their confidence and accuracy in predicting the last digit increasing when conditioned on correct prior digits, seem to be related, and could stem from the view that autoregressive language models are ``exponentially diverging diffusion processes,'' a view that several researchers have argued informally~\cite{lecun2023debate}, and has also recently been more formally proven~\cite{dziri2023faith}. The argument is essentially that if an autoregressive LLM has some non-zero chance of making a mistake, then repeated application of that LLM to generate a long string will cause errors to compound exponentially.

This argument is not fully satisfying, however, for explaining the behavior of LLMs in predicting the last digit. Not only should $p(last\_digit \vert wrong\_intervening\_digits)$ be the same as $p(last\_digit \vert correct\_intervening\_digits)$ due to the computation involved (the last digit not depending on any other digits of the answer at all), but the fact that LLMs are more correct and more confident when conditioned on correct digits rather than wrong digits means that LLMs are able to internally distinguish between the two states, despite not being able to generate the entire correct string in the first place.

This finding may be related to recent results in the hallucination detection literature, where it has been noted that the internal states of LLMs can be used to detect when the conditioning text, including its own outputs, are wrong~\cite{azaria2023lying, chen2024inside}. It stands to reason that if the internal states of an LLM differ depending on whether its conditioning is correct or not, then further outputs which are autoregressively generated based on these internal states may also differ. In other words, while previous results show that LLMs may experience exponentially compounding errors, our finding suggests this may occur not only due to faulty reasoning when using incorrect intermediate steps, but also when the LLM ``realizes'' that it had generated incorrect output, and then ``believes'' that its task is to continue to do so. While out of the scope of this paper, we are interested in further study of this property in particular, and its potential implications.

\section{Conclusion}

Here we present findings on the application of LLMs to arithmetic tasks, seen through the lens of Monte Carlo Dropout. We found that the abilities of what LLMs can do in practice, versus what the theory dictates should be possible for LLMs to do, can be reversed in several cases.
In particular, we found that Llama 2 and Mistral could confidently and correctly output the first digit of the result of $n$-digit by $m$-digit multiplication tasks despite most likely being unable to in the general case, whereas they struggled with outputting the last digit either correctly or confidently, a task which should be easily learnable. We also found that accuracy and confidence in outputting the last digit increases when the prior digits are correct, and we believe that this finding is related to, and could have implications for, recent results in hallucination detection.

\section{Limitations}

MC Dropout is a technique that is only applicable when neural network weights are available and the neural network was trained with dropout. These restrictions limit the number of language models that can be analyzed with the techniques in this paper significantly, and crucially, state of the art language models such as GPT-4~\cite{OpenAIGPT42023}, Gemini~\cite{team2023gemini}, and Claude~\cite{anthropic2023claude} cannot be analyzed in this way by researchers outside of OpenAI, Google, and Anthropic respectively. Such limitations make clear the need for researchers to have access to language models with open weights.

As we have restricted our analysis to Llama 2 and Mistral (which share similar architectures), it is possible that our findings do not generalize to other large language models, but given the very small number of existing language models that can be analyzed in this way, it will be difficult to gauge the generality of our findings until more language models which were trained with dropout and have open weights are released.

\bibliography{references}

\begin{thebibliography}{26}
\providecommand{\natexlab}[1]{#1}

\bibitem[{{Anthropic}(2023)}]{anthropic2023claude}
{Anthropic}. 2023.
\newblock {Model Card and Evaluations for Claude Models}.

\bibitem[{Ashukha et~al.(2020)Ashukha, Lyzhov, Molchanov, and Vetrov}]{ashukha2020pitfalls}
Arsenii Ashukha, Alexander Lyzhov, Dmitry Molchanov, and Dmitry Vetrov. 2020.
\newblock {Pitfalls of in-domain uncertainty estimation and ensembling in deep learning}.
\newblock In \emph{8th International Conference on Learning Representations, ICLR 2020}.

\bibitem[{Azaria and Mitchell(2023)}]{azaria2023lying}
Amos Azaria and Tom Mitchell. 2023.
\newblock \href {https://doi.org/10.18653/v1/2023.findings-emnlp.68} {{The Internal State of an LLM Knows When It's Lying}}.
\newblock In \emph{Findings of the Association for Computational Linguistics: EMNLP 2023}.

\bibitem[{Brown et~al.(2020)Brown, Mann, Ryder, Subbiah, Kaplan, Dhariwal, Neelakantan, Shyam, Sastry, Askell, Agarwal, Herbert-Voss, Krueger, Henighan, Child, Ramesh, Ziegler, Wu, Winter, Hesse, Chen, Sigler, Litwin, Gray, Chess, Clark, Berner, McCandlish, Radford, Sutskever, and Amodei}]{brown2020language}
Tom~B. Brown, Benjamin Mann, Nick Ryder, Melanie Subbiah, Jared Kaplan, Prafulla Dhariwal, Arvind Neelakantan, Pranav Shyam, Girish Sastry, Amanda Askell, Sandhini Agarwal, Ariel Herbert-Voss, Gretchen Krueger, Tom Henighan, Rewon Child, Aditya Ramesh, Daniel~M. Ziegler, Jeffrey Wu, Clemens Winter, Christopher Hesse, Mark Chen, Eric Sigler, Mateusz Litwin, Scott Gray, Benjamin Chess, Jack Clark, Christopher Berner, Sam McCandlish, Alec Radford, Ilya Sutskever, and Dario Amodei. 2020.
\newblock {Language models are few-shot learners}.
\newblock In \emph{Advances in neural information processing systems 33}, pages 1877--1901.

\bibitem[{Chen et~al.(2024)Chen, Liu, Chen, Gu, Wu, Tao, Fu, and Ye}]{chen2024inside}
Chao Chen, Kai Liu, Ze~Chen, Yi~Gu, Yue Wu, Mingyuan Tao, Zhihang Fu, and Jieping Ye. 2024.
\newblock {INSIDE: LLMs' Internal States Retain the Power of Hallucination Detection}.
\newblock In \emph{The Twelfth International Conference on Learning Representations}.

\bibitem[{Dziri et~al.(2023)Dziri, Lu, Sclar, Li, Jiang, Lin, West, Bhagavatula, Le~Bras, Hwang, Sanyal, Welleck, Ren, Ettinger, Harchaoui, and Choi}]{dziri2023faith}
Nouha Dziri, Ximing Lu, Melanie Sclar, Lorraine Li, Liwei Jiang, Bill~Yuchen Lin, Peter West, Chandra Bhagavatula, Ronan Le~Bras, Jena~D Hwang, Soumya Sanyal, Sean Welleck, Xiang Ren, Allyson Ettinger, Zaid Harchaoui, and Yejin Choi. 2023.
\newblock {Faith and Fate: Limits of Transformers on Compositionality}.
\newblock In \emph{Thirty-seventh Conference on Neural Information Processing Systems}.

\bibitem[{Gal and Ghahramani(2016)}]{Gal2015}
Yarin Gal and Zoubin Ghahramani. 2016.
\newblock {Dropout as a Bayesian approximation: Representing model uncertainty in deep learning}.
\newblock In \emph{33rd International Conference on Machine Learning, ICML 2016}, volume~3, pages 1651--1660.

\bibitem[{{Gemini Team} et~al.(2023){Gemini Team}, Anil, Borgeaud, Wu, Alayrac, Yu, Soricut, Schalkwyk, Dai, Hauth, and {others}}]{team2023gemini}
{Gemini Team}, Rohan Anil, Sebastian Borgeaud, Yonghui Wu, Jean-Baptiste Alayrac, Jiahui Yu, Radu Soricut, Johan Schalkwyk, Andrew~M Dai, Anja Hauth, and {others}. 2023.
\newblock {Gemini: a family of highly capable multimodal models}.
\newblock \emph{arXiv preprint arXiv:2312.11805}.

\bibitem[{{Gemma Team}(2024)}]{gemma2024gemma}
{Gemma Team}. 2024.
\newblock {Gemma: Open models based on gemini research and technology}.
\newblock \emph{arXiv preprint arXiv:2403.08295}.

\bibitem[{Hendrycks et~al.(2021)Hendrycks, Burns, Basart, Zou, Mazeika, Song, and Steinhardt}]{Hendrycks2021MMLU}
Dan Hendrycks, Collin Burns, Steven Basart, Andy Zou, Mantas Mazeika, Dawn Song, and Jacob Steinhardt. 2021.
\newblock {Measuring Massive Multitask Language Understanding}.
\newblock \emph{Proceedings of the International Conference on Learning Representations (ICLR)}.

\bibitem[{Jiang et~al.(2023)Jiang, Sablayrolles, Mensch, Bamford, Singh~Chaplot, de~las Casas, Bressand, Lengyel, Lample, Saulnier, Renard~Lavaud, Lachaux, Stock, Le~Scao, Lavril, Wang, Lacroix, and El~Sayed}]{jiang2023mistral}
Albert~Q Jiang, Alexandre Sablayrolles, Arthur Mensch, Chris Bamford, Devendra Singh~Chaplot, Diego de~las Casas, Florian Bressand, Gianna Lengyel, Guillaume Lample, Lucile Saulnier, Lélio Renard~Lavaud, Marie-Anne Lachaux, Pierre Stock, Teven Le~Scao, Thibaut Lavril, Thomas Wang, Timothée Lacroix, and William El~Sayed. 2023.
\newblock {Mistral 7B}.
\newblock \emph{arXiv preprint arXiv:2310.06825}.

\bibitem[{Kojima et~al.(2022)Kojima, Gu, Reid, Matsuo, and Iwasawa}]{kojima2022large}
Takeshi Kojima, Shixiang~Shane Gu, Machel Reid, Yutaka Matsuo, and Yusuke Iwasawa. 2022.
\newblock {Large language models are zero-shot reasoners}.
\newblock \emph{Advances in neural information processing systems}, 35:22199--22213.

\bibitem[{LeCun et~al.(2023)LeCun, Lake, Browning, Chalmers, Pavlick, and Lupyan}]{lecun2023debate}
Yann LeCun, Brenden Lake, Jacob Browning, David Chalmers, Ellie Pavlick, and Gary Lupyan. 2023.
\newblock \href {https://youtu.be/x10964w00zk?si=EbXqIwg_ilD6JaC8} {{Debate: Do language models need sensory grounding for meaning and understanding?}}

\bibitem[{Liu et~al.(2023)Liu, Kian, and Low}]{liu2023goat}
Tiedong Liu, Bryan Kian, and Hsiang Low. 2023.
\newblock {Goat: Fine-tuned LLaMA Outperforms GPT-4 on Arithmetic Tasks}.
\newblock \emph{arXiv preprint arXiv:2305.14201}.

\bibitem[{Mo and Xin(2023)}]{mo2023tout}
Shentong Mo and Miao Xin. 2023.
\newblock {Tree of Uncertain Thoughts Reasoning for Large Language Models}.
\newblock \emph{arXiv preprint arXiv:2309.07694}.

\bibitem[{Morris et~al.(2024)Morris, Sohl-Dickstein, Fiedel, Warkentin, Dafoe, Faust, Farabet, and Legg}]{morris2023agilevels}
Meredith~Ringel Morris, Jascha Sohl-Dickstein, Noah Fiedel, Tris Warkentin, Allan Dafoe, Aleksandra Faust, Clement Farabet, and Shane Legg. 2024.
\newblock {Levels of AGI: Operationalizing Progress on the Path to AGI}.
\newblock \emph{arXiv preprint arXiv:2311.02462}.

\bibitem[{Nogueira et~al.(2021)Nogueira, Jiang, Lin, and Cheriton}]{nogueira2021arithmetic}
Rodrigo Nogueira, Zhiying Jiang, Jimmy Lin, and David~R Cheriton. 2021.
\newblock \href {https://github.com/castorini/transformers-arithmetic} {{Investigating the Limitations of Transformers with Simple Arithmetic Tasks}}.
\newblock Technical report.

\bibitem[{{OpenAI}(2023)}]{OpenAIGPT42023}
{OpenAI}. 2023.
\newblock {GPT-4 Technical Report}.
\newblock Technical report.

\bibitem[{Ovadia et~al.(2019)Ovadia, Fertig, Ren, Nado, Sculley, Nowozin, Dillon, Lakshminarayanan, and Snoek}]{ovadia2019uncertainty}
Yaniv Ovadia, Emily Fertig, Jie Ren, Zachary Nado, D.~Sculley, Sebastian Nowozin, Joshua~V. Dillon, Balaji Lakshminarayanan, and Jasper Snoek. 2019.
\newblock {Can you trust your model's uncertainty? evaluating predictive uncertainty under dataset shift}.
\newblock In \emph{Advances in Neural Information Processing Systems}, volume~32.

\bibitem[{Saharia et~al.(2022)Saharia, Chan, Saxena, Li, Whang, Denton, Ghasemipour, Ayan, Mahdavi, Gontijo-Lopes, Salimans, Ho, Fleet, and Norouzi}]{saharia2022imagen}
Chitwan Saharia, William Chan, Saurabh Saxena, Lala Li, Jay Whang, Emily Denton, Seyed Kamyar~Seyed Ghasemipour, Burcu~Karagol Ayan, S.~Sara Mahdavi, Raphael Gontijo-Lopes, Tim Salimans, Jonathan Ho, David~J. Fleet, and Mohammad Norouzi. 2022.
\newblock {Photorealistic Text-to-Image Diffusion Models with Deep Language Understanding}.
\newblock In \emph{Advances in Neural Information Processing Systems}, volume~35.

\bibitem[{Schick et~al.(2024)Schick, Dwivedi-Yu, Dess{\`{i}}, Raileanu, Lomeli, Hambro, Zettlemoyer, Cancedda, and Scialom}]{schick2024toolformer}
Timo Schick, Jane Dwivedi-Yu, Roberto Dess{\`{i}}, Roberta Raileanu, Maria Lomeli, Eric Hambro, Luke Zettlemoyer, Nicola Cancedda, and Thomas Scialom. 2024.
\newblock {Toolformer: Language models can teach themselves to use tools}.
\newblock \emph{Advances in Neural Information Processing Systems}, 36.

\bibitem[{Shelmanov et~al.(2021)Shelmanov, Tsymbalov, Puzyrev, Fedyanin, Panchenko, and Panov}]{shelmanov2021certain}
Artem Shelmanov, Evgenii Tsymbalov, Dmitri Puzyrev, Kirill Fedyanin, Alexander Panchenko, and Maxim Panov. 2021.
\newblock \href {https://doi.org/10.18653/v1/2021.eacl-main.157} {{How certain is your transformer?}}
\newblock In \emph{EACL 2021 - 16th Conference of the European Chapter of the Association for Computational Linguistics, Proceedings of the Conference}.

\bibitem[{Touvron et~al.(2023)Touvron, Martin, Stone, Albert, Almahairi, Babaei, Bashlykov, Batra, Bhargava, Bhosale, and {others}}]{touvron2023llama}
Hugo Touvron, Louis Martin, Kevin Stone, Peter Albert, Amjad Almahairi, Yasmine Babaei, Nikolay Bashlykov, Soumya Batra, Prajjwal Bhargava, Shruti Bhosale, and {others}. 2023.
\newblock {Llama 2: Open foundation and fine-tuned chat models}.
\newblock \emph{arXiv preprint arXiv:2307.09288}.

\bibitem[{Wang et~al.(2023)Wang, Xie, Jiang, Mandlekar, Xiao, Zhu, Fan, and Anandkumar}]{wang2023voyager}
Guanzhi Wang, Yuqi Xie, Yunfan Jiang, Ajay Mandlekar, Chaowei Xiao, Yuke Zhu, Linxi Fan, and Anima Anandkumar. 2023.
\newblock {Voyager: An Open-Ended Embodied Agent with Large Language Models}.
\newblock \emph{arXiv preprint arXiv:2305.16291}.

\bibitem[{Wies et~al.(2022)Wies, Levine, and Shashua}]{wies2022sub}
Noam Wies, Yoav Levine, and Amnon Shashua. 2022.
\newblock {Sub-Task Decomposition Enables Learning in Sequence to Sequence Tasks}.
\newblock In \emph{The Eleventh International Conference on Learning Representations}.

\bibitem[{Wolf et~al.(2019)Wolf, Debut, Sanh, Chaumond, Delangue, Moi, Cistac, Rault, Louf, Funtowicz, and {others}}]{wolf2019huggingface}
Thomas Wolf, Lysandre Debut, Victor Sanh, Julien Chaumond, Clement Delangue, Anthony Moi, Pierric Cistac, Tim Rault, Rémi Louf, Morgan Funtowicz, and {others}. 2019.
\newblock {Huggingface's transformers: State-of-the-art natural language processing}.
\newblock \emph{arXiv preprint arXiv:1910.03771}.

\end{thebibliography}

\appendix

\section{Prompt Format and Hyperparameters}
\label{sec:appendixHyperparam}

The exact prompt used in Sections~\ref{sec:unconditional} and \ref{sec:conditional} is ``$111*472=52392$. $362*194=70228$. $\{math\_question\}=\{given\_str\}$'' where $math\_question$ is the multiplication task, and $given\_str$ is the empty string in Section~\ref{sec:unconditional} and all but the last digit of the correct answer in Section~\ref{sec:conditional}. In Section~\ref{sec:many} the prompts are randomly generated 2-shot $n$-digit by $m$-digit multiplication examples in the same format.

We set the dropout rate to be $0.1$, which is the dropout rate commonly used in GPT applications, and appears to be the dropout rate used to train Llama 2 and Mistral. All sampling from LLMs is done deterministically other than the stochasticity induced by dropout (i.e., we take argmax over logits). We collect $100$ samples for each output.

\section{Tokenization}
\label{sec:appendixTokenization}

Both the Llama 2 and Mistral tokenizers have one single token for each digit, $0$ to $9$, and no digits appear in any tokens other than these. This property has been shown to be necessary to consistently perform even simple addition tasks~\cite{nogueira2021arithmetic}.

\section{Dropout}
\label{sec:appendixDropout}

The use of MC Dropout to model uncertainty in neural networks requires, as a prerequisite, that the neural networks were trained with dropout. As we do not know the exact training details of Llama 2 or Mistral, we cannot be fully assured that they used dropout in training. We do, however, have very strong reason to believe that they did use dropout during training, due to the fact that both of these models still output reasonable text when dropout is turned on. Conversely, the Gemma~\cite{gemma2024gemma} HuggingFace code also has dropout, but when dropout is turned on even to only 10\%, the model outputs are entirely nonsensical (when attempting these experiments with Gemma, we do not even get numbers as output when dropout is turned on, but do get reasonable output with dropout turned off). The sort of robustness to neurons being dropped out that can be seen in Llama 2 and Mistral only occurs in models that were actually trained with dropout, and thus we can be fairly confident that the use of MC Dropout here is appropriate.

\end{document}